\documentclass{article}

\usepackage[nonatbib, main, final]{neurips}

\usepackage{xcolor}         
\definecolor{blue}{RGB}{3, 100, 201}
\definecolor{green}{RGB}{128, 195, 27}
\definecolor{orange}{RGB}{255, 128, 1}

\usepackage{amsfonts}
\usepackage{amsmath}
\usepackage{array}
\usepackage{booktabs}
\usepackage{caption}
\usepackage{enumitem}
\usepackage{graphicx}
\usepackage[misc]{ifsym}
\usepackage{makecell}
\usepackage{multirow}
\usepackage{needspace}
\usepackage{pifont}
\usepackage{soul}
\usepackage{tabularx}
\usepackage{times}
\usepackage{url}
\usepackage{wrapfig}
\usepackage{xspace}
\usepackage[
  bookmarks=false,
  pagebackref=false,
  breaklinks=true,
  colorlinks=true,
  citecolor=blue,
  linkcolor=blue,
]{hyperref}

\newif\ifanonymous

\newcommand{\eg}{\textit{e.g.}}
\newcommand{\ie}{\textit{i.e.}}

\newcommand{\cmark}{\textcolor{green}{\ding{51}}}%
\newcommand{\xmark}{\textcolor{red}{\ding{55}}}%
\newcommand{\dwar}{$\downarrow$}
\newcommand{\upar}{$\uparrow$}

\newcommand{\OM}{\textcolor{blue}{Dynamic}\textcolor{green}{VLA}\xspace}
\newcommand{\OMplain}{DynamicVLA\xspace}
\newcommand{\OMlogo}{\textcolor{blue}{ynamic}\textcolor{green}{VLA}}
\newcommand{\OMsize}{0.4B\xspace}
\newcommand{\OD}{Dynamic Object Manipulation\xspace}
\newcommand{\ODslug}{DOM\xspace}

\newcolumntype{Y}{>{\centering\arraybackslash}X}

\renewenvironment{abstract}{%
  \vskip 0.15in%
  \centerline{\large\bf Abstract}%
  \vspace{-1ex}%
  \begin{quote}%
}{%
  \par\end{quote}\vskip 1ex%
}

\title{%
\makebox[6.25 mm][l]{%
  \raisebox{-0.8 mm}{\includegraphics[height=6 mm]{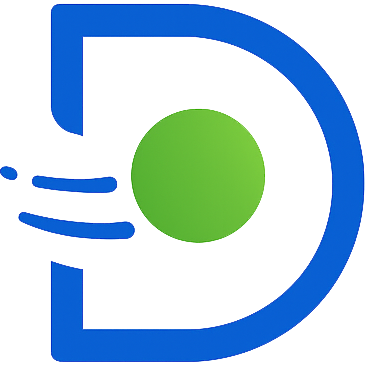}}%
}%
\textbf{\OMlogo}: A Vision-Language-Action\\Model for Dynamic Object Manipulation}
\author{%
  \hspace*{-8 mm}
  Haozhe Xie\textsuperscript{*}\hspace{1 mm}
  Beichen Wen\textsuperscript{*}\hspace{1 mm}
  Jiarui Zheng\hspace{1 mm}
  Zhaoxi Chen\hspace{1 mm}
  Fangzhou Hong\hspace{1 mm}
  Haiwen Diao\hspace{1 mm}
  Ziwei Liu~\textsuperscript{\Letter}\\
  S-Lab, Nanyang Technological University\\
  \url{https://haozhexie.com/project/dynamic-vla}
}

\begin{document}

\makeatletter
\let\@oldmaketitle\@maketitle
\renewcommand{\@maketitle}{%
  \@oldmaketitle
  \begin{center}
    \vspace{-8 mm}
    \includegraphics[width=\textwidth]{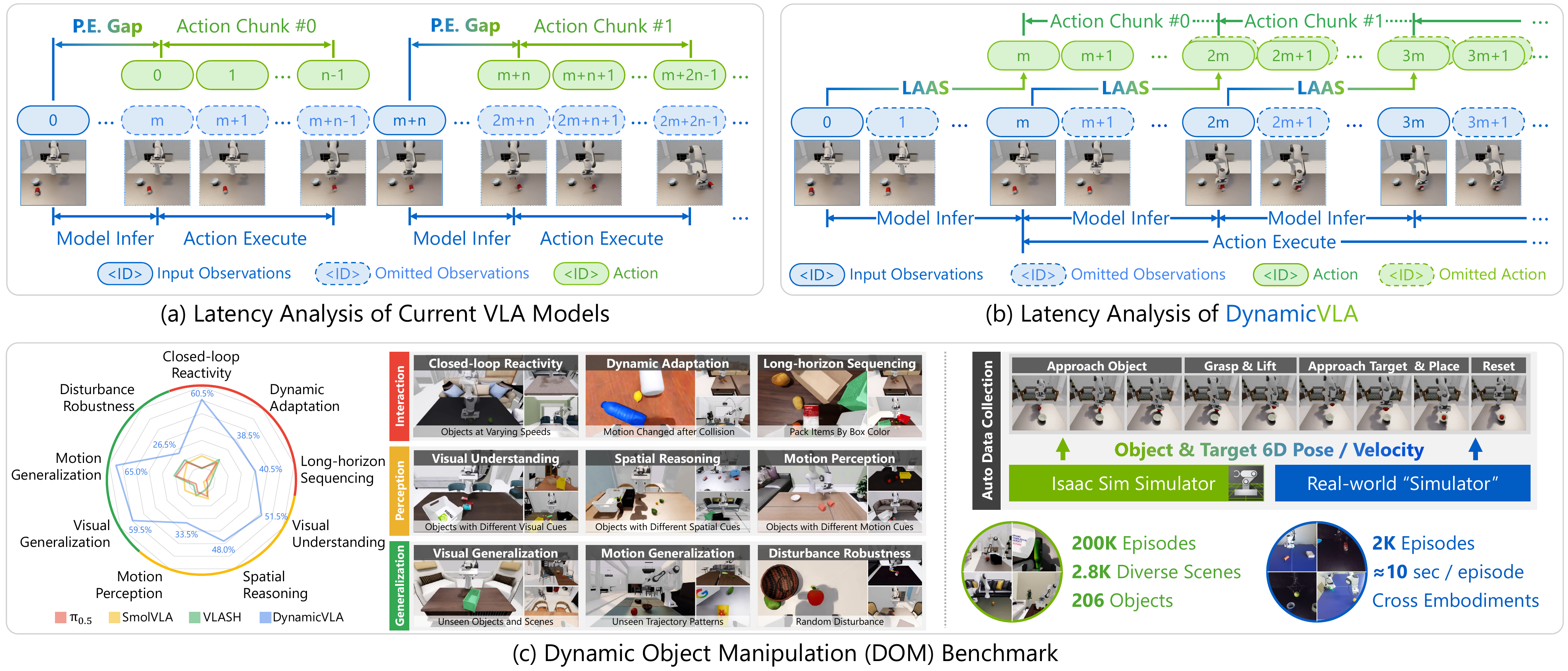}
    \captionof{figure}{
    \textbf{(a)} Current VLA models suffer from inter-chunk waiting and perception--execution (P.E.) mismatch, leading to delayed or misaligned reactions to moving objects.
    \textbf{(b)} \OM uses a continuous inference schedule to maintain a non-blocking action stream, while LAAS discards latency-invalid prefixes and executes only temporally valid suffixes to preserve action-time alignment under dynamic object motion.
    \textbf{(c)} The \OD (\ODslug) benchmark features 2.8K scenes and 206 objects for evaluating Perception, Interaction, and Generalization, while its data collection pipeline enables gathering of 200K synthetic and 2K real-world episodes.}
    \label{fig:teaser}
    \vspace{-1 mm}
  \end{center}
}
\let\@oldnoticestring\@noticestring
\renewcommand{\@noticestring}{%
  \noindent\rule{12pc}{0.4pt}\par
  \vspace{2pt}%
  \textsuperscript{*}~Equal contribution. \hspace{2 mm} \textsuperscript{\Letter}~Corresponding author.\par
  \vspace{3pt}%
  \@oldnoticestring
}
\makeatother

\maketitle
\enlargethispage{\baselineskip}

\begin{abstract}
Manipulating dynamic objects remains an open challenge for Vision-Language-Action (VLA) models. 
Although recent VLAs generalize well in static manipulation, dynamic scenes introduce a latency-induced perception–execution mismatch: object states continue to evolve during inference, making actions predicted from past observations stale at execution time.
We present \OM, a latency-aware VLA model for dynamic object manipulation. 
It combines a compact \OMsize architecture and convolutional vision encoder for efficient multimodal inference with a continuous inference schedule that overlaps reasoning and execution for non-blocking control. 
Latency-aware Action Streaming then discards latency-invalid action prefixes and executes only the temporally valid suffix of each predicted chunk, preserving action-time alignment under dynamic object motion.
To fill the missing foundation of dynamic manipulation data, we introduce the \OD (\ODslug) benchmark, built with an automated collection pipeline that gathers 200K synthetic episodes across 2.8K scenes and 206 objects, and enables fast collection of 2K real-world episodes without teleoperation.
Extensive evaluations in simulation and on real robots show that \OMplain improves dynamic manipulation success under changing object motion, perception-heavy instructions, and unseen motion patterns.
\end{abstract}

\section{Introduction}

Dynamic object manipulation is a fundamental yet underexplored frontier in robotics. 
Real-world interaction often involves objects in motion, such as handing, repositioning, or stabilizing items, requiring robots to perceive, predict, and act under changing conditions.
Even minor latency may cause failure, making dynamic manipulation substantially more challenging than static grasping~\cite{DBLP:preprint/arxiv/2510-10903}.

Robots have long been evaluated on moving-target tasks such as throwing~\cite{DBLP:journals/trob/ZengSLRF20}, soccer~\cite{DBLP:conf/agents/KitanoAKNO97}, and table tennis~\cite{DBLP:preprint/arxiv/2604-01158, DBLP:journals/nature/PeterMJNNSNYPY26}, but typically rely on reactive control and handcrafted perception in structured settings. 
Recent manipulation methods~\cite{DBLP:journals/ral/ZhangWC25, DBLP:preprint/arxiv/2508-14042} handle moving objects, yet remain limited to predictable, conveyor-belt–like motion. 
Concurrent VLAs~\cite{DBLP:preprint/github/rdt2, DBLP:preprint/arxiv/2510-26742, DBLP:preprint/arxiv/2512-01031} demonstrate real-time interaction with fast-moving targets, but mainly in striking or interception-style tasks with coarse contact and large spatial margins. 
Dynamic object manipulation instead requires precise 6DoF grasping and placement under uncertain motion, making tight perception--action alignment essential and still underexplored.

VLA models~\cite{DBLP:preprint/arxiv/2507-01925} have achieved strong results in static manipulation, where object states remain nearly unchanged during inference and latency is often tolerable. 
Even large 3B--7B VLAs~\cite{DBLP:conf/corl/KimPKXBNRFSVKBTSLF25, DBLP:conf/rss/BlackBDEEFFGH25} perform well despite slow inference, while recent designs~\cite{DBLP:preprint/arxiv/2506-01844, DBLP:preprint/arxiv/2509-09372} further improve efficiency. 
Asynchronous action-chunking methods~\cite{DBLP:conf/rss/ZhaoKLF23, DBLP:conf/nips/BlackGL25} keep action chunks running and improve chunk-to-chunk continuity, but they do not guarantee temporal alignment between predicted actions and evolving object states.
As shown in Figure~\ref{fig:teaser}\hyperref[fig:teaser]{a}, object states keep evolving during inference, making early predicted actions temporally invalid before execution. 
This latency-induced perception--execution gap is mostly hidden in static benchmarks but becomes critical for dynamic objects.

To address these issues, we propose \OM, a latency-aware VLA model for dynamic object manipulation.
It uses a compact \OMsize-parameter architecture with a convolutional vision encoder for efficient spatial compression and visual encoding.
During execution, we use a continuous inference schedule that overlaps prediction and control, avoiding inter-chunk waiting and letting newer chunks supersede older ones at overlapping steps.
Yet continuous inference maintains an uninterrupted action stream; it does not correct the temporal-index shift caused by inference latency.
Since the earliest actions in each chunk are indexed to execution times that may have already elapsed, Latency-aware Action Streaming discards these latency-invalid prefixes and executes only the temporally valid suffix of each chunk, preserving action-time alignment under dynamic object motion.

A key barrier to developing and evaluating VLA policies for dynamic scenarios is the lack of large-scale data and standardized protocols. 
Existing robotic datasets~\cite{DBLP:conf/nips/LiuZGFLZS23, DBLP:conf/icra/ONeillRMGPLPGMJ24} focus mostly on static scenes, while teleoperation fails when objects move faster than humans can reliably track. 
To close this gap, we introduce \OD (\ODslug), a benchmark with an automated data-collection pipeline validated on Franka Emika Panda and AgileX PiPER. 
In simulation, Isaac Sim~\cite{DBLP:preprint/arxiv/2511-04831} provides real-time 6D object poses and velocities to drive an automatic demonstrator, yielding 200K episodes across 2.8K simulation-ready 3D scenes and 206 objects. 
For real-world collection, a ``real-world'' simulator exposes the same state interface using estimated object position and velocity, enabling the same demonstrator to collect 2K episodes without teleoperation.

Our contributions are summarized as follows.
\begin{itemize}[nosep, leftmargin=*]
\item We identify latency-induced perception--execution gap as a failure mode for VLA policies in dynamic manipulation, and present \OM, a latency-aware VLA model that combines multimodal inference with non-blocking inference and action-time-aligned streaming.

\item We introduce \OD (\ODslug), a large-scale dynamic manipulation benchmark with an automated data-collection pipeline, comprising 200K synthetic episodes, 2.8K scenes, 206 objects, and 2K real-world episodes across multiple embodiments.

\item We conduct extensive simulation and real-robot evaluations on \ODslug and 16 real-robot tasks, showing that \OMplain improves dynamic manipulation under changing object motion, perception-heavy instructions, and unseen objects, scenes, and motion regimes.
\end{itemize}
\section{Related Work}

\noindent \textbf{Vision-Language-Action Models.}
Inspired by the success of 
Large Language Models (LLMs)~\cite{DBLP:preprint/arxiv/2303-08774, DBLP:preprint/arxiv/2407-21783, DBLP:journals/nature/GuoYZSWZXZMB25} and 
Vision-Language Models (VLMs)~\cite{DBLP:conf/nips/LiuLWL23a, DBLP:conf/cvpr/LiuLLL24, DBLP:preprint/arxiv/2502-13923}, Vision-Language-Action (VLA) models extend VLMs with action generation.
Transformer-based methods~\cite{DBLP:conf/rss/ZhaoKLF23, DBLP:conf/rss/BrohanBCCDFGHHH23} use Transformers to model state-action-reward sequences. 
LLM/VLM-based methods~\cite{DBLP:conf/corl/ZitkovichYXXXXW23, DBLP:conf/corl/KimPKXBNRFSVKBTSLF25} treat VLA tasks as sequence-to-sequence problems for action generation. 
Diffusion-based methods~\cite{DBLP:conf/rss/ChiFDXCBS23, DBLP:preprint/arxiv/2502-19645} model policies as denoising diffusion models. 
LLM and diffusion model methods~\cite{DBLP:conf/rss/GhoshWPBMDHK0LT24, DBLP:conf/rss/BlackBDEEFFGH25} combine LLMs for representation and diffusion models for action generation. 
Video generation with inverse kinematics methods~\cite{DBLP:conf/iclr/YangDGTKSA24, DBLP:conf/iclr/WuJCCXLLLK24, DBLP:conf/cvpr/RenWG0KFJ25} generate motion sequences and convert them into actions.
However, existing VLA models are often too slow for precise or rapid execution.

\noindent \textbf{Robot Learning Datasets.}
%
%
%
Real-world datasets~\cite{DBLP:conf/corl/MandlekarZGBSTG18, DBLP:conf/corl/WalkeBZVZHHMKDL23, DBLP:conf/icra/ONeillRMGPLPGMJ24, DBLP:conf/rss/WuHLCJYLZXY25} provide high-fidelity interactions but are costly and hard to scale, while simulated datasets~\cite{DBLP:journals/ral/MeesHRB22, DBLP:conf/nips/LiuZGFLZS23, DBLP:preprint/arxiv/2403-09227, DBLP:conf/corl/LiHGMPWFLSKL0F024, DBLP:conf/cvpr/MuCCPLGLYZXLXDL25} offer scalability yet suffer from the sim-to-real gap.
Most benchmarks focus on simple tabletop manipulation (\eg, pick-and-place, pushing) with limited task diversity, though recent work explores long-horizon~\cite{DBLP:conf/nips/LiuZGFLZS23, DBLP:preprint/arxiv/2412-18194}, language-conditioned~\cite{DBLP:conf/nips/ZhengCJW22, DBLP:preprint/arxiv/2210-03094}, and tactile-rich~\cite{DBLP:conf/corl/0028K0GAMS22, DBLP:journals/trob/AkinolaXCFN25} settings.
Generative models~\cite{DBLP:conf/rss/NasirianyMZPLJMZ24, DBLP:conf/icml/WangXCWWFEHG24, DBLP:conf/iclr/YangDGTKSA24, DBLP:preprint/arxiv/2505-05474} introduce interactive elements but remain constrained by artifacts, low frame rates, and memory.
Despite progress in standardization and multi-embodiment learning, current datasets lack dynamic objects, limiting applicability to environments with independent motion.

\noindent \textbf{Robot Dynamic Manipulation.}
Robots have long handled moving targets in specialized settings such as table tennis~\cite{DBLP:preprint/arxiv/2604-01158, DBLP:journals/nature/PeterMJNNSNYPY26}, but mostly through specialized systems.
Recent moving-object manipulation methods mainly consider structured or conveyor-like motion~\cite{DBLP:journals/ral/ZhangWC25, DBLP:preprint/arxiv/2508-14042}, while concurrent VLA efforts have begun to explore real-time or dynamic interaction~\cite{DBLP:preprint/github/rdt2, DBLP:preprint/arxiv/2510-26742, DBLP:preprint/arxiv/2512-01031}.
However, general dynamic object manipulation with fine contact requirements, evaluated broadly in both simulation and the real world, remains less studied.


\section{The \OMplain Model}
\label{sec:method}

\subsection{Problem Formulation}

We study dynamic object manipulation, where object states evolve continuously during perception, reasoning, and execution. 
At time step $t$, the VLA policy $\pi$ receives a temporal window of visual observations $\mathbf{O}_t = \{\mathbf{o}_{t - k}, \dots, \mathbf{o}_t\}$, a language instruction $\mathbf{L}_t$, and its proprioceptive state $\mathbf{P}_t$, and predicts an action sequence $\mathbf{A}_t = \{\mathbf{a}_t, \dots, \mathbf{a}_{t + n}\}$,
\ie, $\mathbf{A}_t = \pi(\mathbf{O}_t, \mathbf{L}_t, \mathbf{P}_t)$.

The physical environment includes a latent object state $\mathbf{s}_t$, describing the object’s 6D pose and motion.
Crucially, object motion does not pause during inference: while the model reasons over $\mathbf{O}_t$, the object transitions from $\mathbf{s}_t$ to $\mathbf{s}_{t + m}$, where $m$ denotes inference latency, leading to potential misalignment between perception and execution.

\begin{wrapfigure}{r}{0.48\linewidth}
  \vspace{-12.5 mm}
  \includegraphics[width=\linewidth]{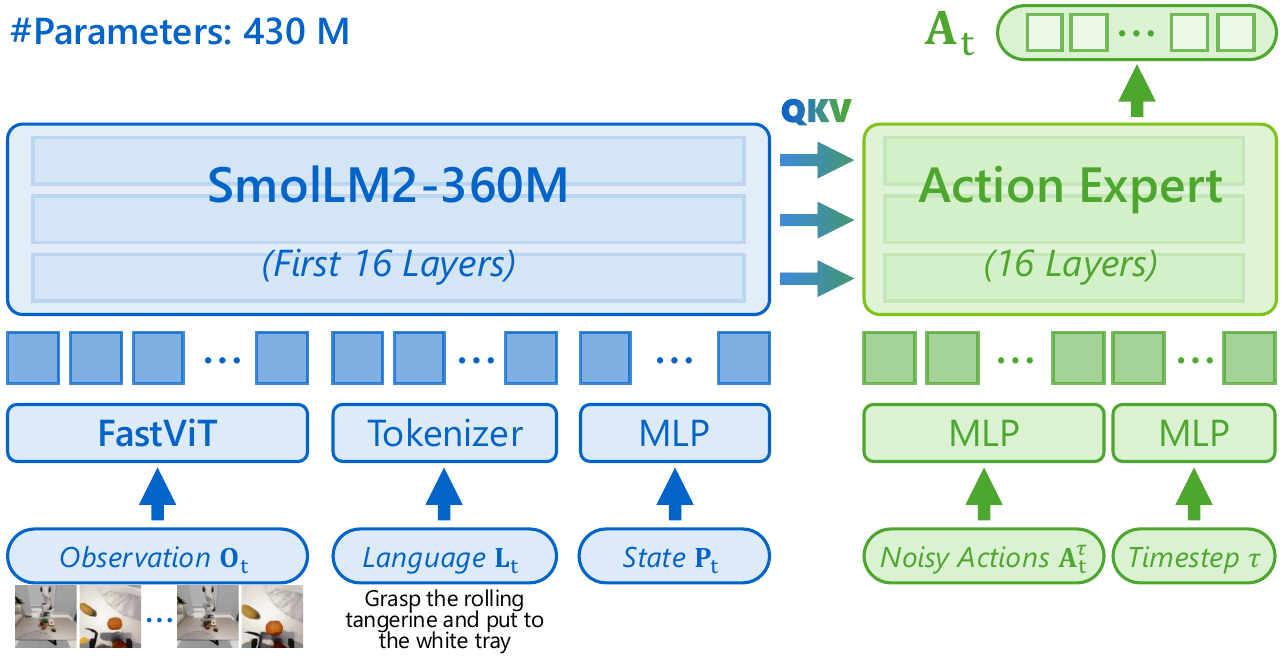}
  \caption{\textbf{The \OMplain architecture.} Multimodal features condition a diffusion-based action expert for action-chunk prediction.}
  \vspace{-4 mm}
  \label{fig:method}
\end{wrapfigure}

\subsection{The \OMplain Architecture}
\label{sec:vla-model}

Since inference latency directly affects whether a robot can react to moving objects, we design a \OMsize VLA model for fast and spatially efficient multimodal reasoning, illustrated in Figure~\ref{fig:method}.

\noindent \textbf{Vision–Language Backbone.}
We adopt SmolLM2-360M~\cite{DBLP:preprint/arxiv/2502-02737} as the language backbone, resulting in an overall tiny model size.
Unlike VLMs that rely on transformer-based vision encoders, we employ a convolutional vision encoder, FastViT~\cite{DBLP:conf/iccv/VasuGZTR23}, which performs efficient spatial compression and avoids quadratic token growth when processing multi-frame visual inputs.
Following SmolVLA~\cite{DBLP:preprint/arxiv/2506-01844}, we use the first 16 transformer layers of the language backbone to balance latency and multimodal reasoning capacity.

\noindent \textbf{Diffusion-Based Action Expert.}
The action expert $\mathcal{E}_\theta$ predicts an action chunk $\mathbf{A}_t$ conditioned on the features produced by the VLM backbone.
Following diffusion-style action modeling~\cite{DBLP:conf/iclr/LipmanCBNL23, DBLP:conf/icml/EsserKBEMSLLSBP24}, we instantiate $\mathcal{E}_\theta$ as a conditional Flow Matching Transformer~\cite{DBLP:conf/rss/BlackBDEEFFGH25} and train it using the objective
\begin{equation}
  \ell^\tau (\theta) = \mathbb{E}_{p(\mathbf{A}_t | \mathbf{f}_t), q(\mathbf{A}_t^\tau|\mathbf{A}_t)}\left[\lVert \mathcal{E}_\theta(\mathbf{A}_t^\tau, \mathbf{O}_t) - \mathbf{u}(\mathbf{A}_t^\tau|\mathbf{A}_t) \rVert\right]
  \label{eq:flow-matching-loss}
\end{equation}
where
superscript $\tau \in [0, 1]$ denotes flow matching timesteps.
$q(\mathbf{A}_t^\tau|\mathbf{A}_t) = \mathcal{N}(\tau\mathbf{A}_t, (1 - \tau)\mathbf{I})$.
$\mathbf{f}_t$ represents the VLM features extracted from $\mathbf{O}_t$, and $\mathbf{A}^\tau_t = \tau\mathbf{A}_t + (1 - \tau)\epsilon$, with $\epsilon \sim \mathcal{N}(0, \mathbf{I})$.
Under this objective, $\mathcal{E}_\theta(\mathbf{A}_t^\tau, \mathbf{O}_t)$ learns to match the denoising vector field $\mathbf{u}(\mathbf{A}_t^\tau \mid \mathbf{A}_t) = \epsilon - \mathbf{A}_t$.

\noindent \textbf{Multi-modal Fusion and Projection.}
We use lightweight linear projections to align representations across modules, including
\textbf{1)} embedding robot states into the multimodal feature space,
\textbf{2)} adapting action representations to the diffusion-based action expert, and
\textbf{3)} matching feature dimensions between the VLM backbone and the action expert.

\subsection{Continuous Inference Schedule}
\label{sec:continuous-infer}

At time step $t$, the VLA model predicts an action sequence $\mathbf{A}_t = \left\{\mathbf{a}_t, \dots, \mathbf{a}_{t + n}\right\}$.
In standard chunk-based VLA execution~\cite{DBLP:conf/rss/KimFL25, DBLP:conf/rss/BlackBDEEFFGH25, DBLP:conf/corl/PhysicalIntelligence25}, the next inference call is triggered only after the current chunk is fully consumed, causing inter-chunk waiting when objects continue to move.

To avoid this waiting, we use a greedy asynchronous schedule that launches the next inference call immediately after the previous one finishes. 
Let $m$ denote the inference delay. 
Although this delay may vary across cycles in practice, we use a uniform $m$ for notational simplicity, so new chunks become available at $t,t+m,t+2m,\dots$.
When a new chunk overlaps with actions remaining from an older chunk, the newer prediction replaces the older one at the same execution timestep.

\subsection{Latency-aware Action Streaming}
\label{sec:laas}

When inference starts at time $t$, the VLA model predicts $\mathbf{A}_t$ from observations $\mathbf{O}_t$. 
However, this chunk becomes available at $t + m$, by which time the object state has evolved from $\mathbf{s}_t$ to $\mathbf{s}_{t + m}$.
Consequently, the prefix $\{\mathbf{a}_t,\dots,\mathbf{a}_{t+m-1}\}$ is indexed to execution times that have elapsed before the chunk can be executed. 
Executing these actions at later physical timesteps would re-time commands predicted for past object states to the current dynamic scene, causing a perception--execution gap. 

Latency-aware Action Streaming addresses this temporal-index validity problem by starting execution from $\mathbf{a}_{t+m}$, the first action whose intended execution time has not elapsed. 
It discards the latency-invalid prefix and streams only the valid suffix $\{\mathbf{a}_{t+m}, \dots, \mathbf{a}_{t+n}\}$. 
This rule preserves action-time alignment by ensuring that each executed action corresponds to a timestep that remains valid when the chunk becomes available, which is critical when object states evolve during inference.

\section{The \OD Benchmark}
\label{sec:dom-dataset}

\subsection{Overview}

\OD (\ODslug) is a large-scale benchmark for dynamic object manipulation, addressing the lack of standardized evaluation on moving objects. 
It provides scalable, fully automated data collection in simulation and the real world, yielding 200K synthetic and 2K real-world episodes where teleoperation is impractical due to human reaction limits. 
\ODslug further structures scenarios by interaction, perception, and generalization factors, enabling consistent comparison across algorithms and robot embodiments.

\subsection{Benchmark Dimensions}

As illustrated in Figure~\ref{fig:teaser}\hyperref[fig:teaser]{c}, \ODslug evaluates dynamic manipulation ability across three dimensions:

\noindent \textbf{Interaction.}
This dimension evaluates how a policy responds to evolving object motion.
\emph{1) Closed-loop reactivity}, which measures how quickly the robot adjusts to objects moving at different speeds;
\emph{2) Dynamic adaptation}, where the policy must handle abrupt motion changes such as direction shifts or unexpected disturbances;
\emph{3) Long-horizon sequencing}, which assesses whether the policy maintains coherent behavior over extended interactions and prioritizes actions as motion events unfold.

\noindent \textbf{Perception.}
This dimension evaluates how a policy perceives and grounds visual and linguistic cues in environments.
\emph{1) Visual understanding}, which measures the ability to distinguish objects with similar shapes, textures, or materials;
\emph{2) Spatial reasoning}, which examines whether the policy can infer object positions and relative arrangements in cluttered or changing scenes;
\emph{3) Motion perception}, which assesses how the policy interprets object motion cues such as speed and direction.

\noindent \textbf{Generalization.}
This dimension evaluates how robustly a policy transfers across novel objects, scenes, and motion patterns.
\emph{1) Visual generalization}, which measures adaptation to unseen shapes, appearances, and scene layouts;
\emph{2) Motion generalization}, which assesses whether the policy can handle new speed ranges, altered friction conditions, and trajectory patterns that differ from those observed during training;
\emph{3) Disturbance Robustness}, which tests the ability to maintain stable behavior under external perturbations such as unexpected pushes, collisions, or sensor noise.

\begin{figure*}
  \resizebox{\linewidth}{!}{
    \includegraphics[width=\textwidth]{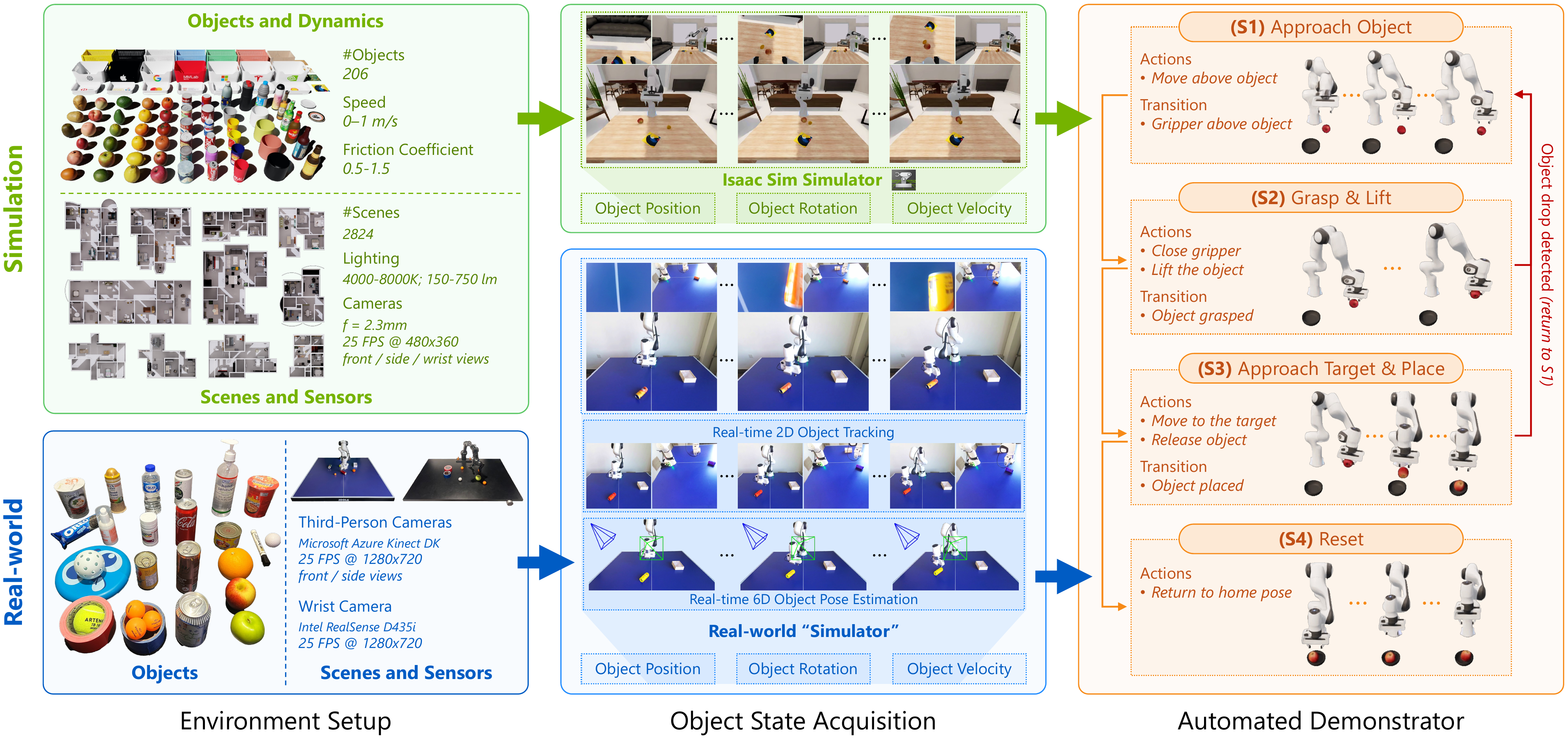}
  }
  \caption{\textbf{Automatic Simulation and Real-world Data Collection.} 
  \emph{Environment Setup:} simulation and real-world settings share objects, scenes, and synchronized cameras.
  \emph{Object State Acquisition:} simulation provides ground-truth 6D object states, while real-world RGB observations are converted into a real-world ``simulator'' interface that enables automatic dynamic-manipulation data collection without teleoperation.}
  \label{fig:data-collection}
\end{figure*}

\subsection{Simulation Data Collection}

Our simulation framework serves two purposes: 
1) scaling dynamic manipulation data for VLA pretraining, and 
2) providing a reproducible benchmark for fair evaluation.
As shown in Figure~\ref{fig:data-collection}, we build a high-throughput Isaac Sim pipeline~\cite{DBLP:preprint/arxiv/2511-04831} that combines scene and object sampling, multi-view RGB rendering, simulator-state acquisition, and automatic demonstration generation.

\noindent \textbf{Objects and Dynamics.}
We use 206 everyday Objaverse objects~\cite{DBLP:conf/cvpr/DeitkeSSWMVSEKF23}, including fruits, vegetables, containers, and household items, with texture augmentation for visual diversity.
Object speeds are sampled from 0--1 m/s (0.15--0.75 m/s for training, with some static objects; see Appendix~\ref{sec:eval-setup}), and friction coefficients from 0.5--1.5.
Multiple objects are placed in the workspace to allow natural interactions during motion.

\noindent \textbf{Scenes and Sensors.}
We use 2.8K 3D-FRONT scenes~\cite{DBLP:conf/iccv/FuC0ZWLZSJZ021}, curated for clean, flat tabletops and realistic, non-self-occluding object placements.
Each scene has three 25-FPS RGB cameras at 480$\times$360 resolution: two third-person views 1 m from the robot and one wrist-mounted camera, with Azure-Kinect-aligned intrinsics.
We randomize illumination with color temperature 4000--8000 K, intensity 150--750 lm, and light positions sampled from $x,y \in [-50,50]$ m and $z \in [10,20]$ m.

\noindent \textbf{Object State Acquisition.}
The simulator maintains ground-truth 6D object states throughout each episode.
Isaac Sim propagates object motion through the physics engine, from which we extract per-object position, rotation, and linear/angular velocity at 25 Hz.
These states form a simulator-style interface for driving the automatic demonstrator during simulation data collection.

\noindent \textbf{Automatic Demonstrator.}
The automatic demonstrator consumes real-time object pose and velocity, together with the target pose, to generate demonstrations through four stages: approach, grasp-and-lift, place, and reset.
It predicts short-horizon object motion during approach, continuously updates the end-effector target, stabilizes residual motion during grasping, and returns the robot to the home pose after placement.
This yields scalable dynamic manipulation trajectories without exposing privileged states, such as object poses and velocities, to learned policies.

\subsection{Real-World Data Collection}

Teleoperation is unreliable for dynamic manipulation because human reaction is too slow for fast-moving objects, while real-world setups lack the ground-truth 6D object states used by simulator pipelines. 
To address both issues, we build a real-world ``simulator'' that estimates simulator-style object states from commodity RGB sensors, enabling teleoperation-free dynamic data collection at $\approx$10 s per episode on both Franka and PiPER.

\noindent \textbf{Environment Setup.}
We use 25 physical household objects spanning containers, food items, bottles, and tools, with multiple objects per episode, including pick/place targets and natural distractors.
The scene is captured by two synchronized third-person RGB cameras (Azure Kinect DK) placed at front and side viewpoints, along with a wrist-mounted RealSense D435i, matching the simulation geometry and supplying synchronized, calibrated RGB streams for state estimation.

\noindent \textbf{Object State Acquisition.}
To replicate the simulator's state interface, we build a real-world ``simulator'' that estimates object position and velocity.
EfficientTAM~\cite{DBLP:preprint/arxiv/2411-18933} supplies per-view object masks from the synchronized third-person cameras, and a geometric triangulation step recovers the 3D centroid.
Linear velocity is estimated by least-squares fitting over a short temporal window, and for curved motion the turning rate of the velocity direction is additionally estimated to extrapolate curved paths, producing a smooth, low-latency state stream that matches the simulator-style interface.
Object orientation and angular velocity are not estimated in the real world: the demonstrator derives grasp orientation from the motion direction and per-object grasp priors, which is sufficient for the rotationally symmetric or top-down-graspable objects used in real-world collection.

\noindent \textbf{Automatic Demonstrator.}
The same automatic demonstrator used in simulation runs unchanged in the real world, consuming estimated object states and target poses to generate trajectories.
These estimated states are used only for data collection and are not exposed to learned policies.
\section{Experiments}

Our experiments evaluate \OMplain on dynamic object manipulation under real-time constraints.
We benchmark \OMplain against VLA baselines across interaction, perception, and generalization challenges, and further analyze how model efficiency, asynchronous execution, and action-time alignment affect performance.
Specifically, we study the following research questions:
\begin{enumerate}[nosep, leftmargin=*]
\item How well can \OMplain interact with fast-moving objects and maintain stable closed-loop behavior over long horizons?
\item How well does \OMplain ground appearance, spatial, and motion cues?
\item How well does \OMplain generalize to unseen objects, scenes, and motion regimes?
\item How do model efficiency, non-blocking execution, and action-time-aligned streaming contribute to performance under inference latency?\end{enumerate}

\subsection{Evaluation Protocols}

\noindent \textbf{Experimental Setup.}
%
\OMplain is evaluated on the \OD (\ODslug) benchmark in simulation and the real world.
Experiments are conducted in Isaac Sim with a Franka Emika Panda arm, and on real-world Franka and AgileX PiPER arms.
In real-world trials, a secondary robot arm induces object motion using predefined launch trajectories, ensuring comparable dynamics across methods.
The evaluated policy controls only the manipulation arm, with access limited to RGB observations, language instructions, and proprioception.
Each real-world task is evaluated over three paired motion--position configurations, with 20 trials per configuration, resulting in 60 trials per task and 960 real-world trials per method.
All methods are evaluated under identical conditions within each environment.

\noindent \textbf{Baselines.}
In simulation, we evaluate ACT~\cite{DBLP:conf/rss/ZhaoKLF23}, Diffusion Policy~\cite{DBLP:conf/rss/ChiFDXCBS23}, OpenVLA-OFT~\cite{DBLP:conf/rss/KimFL25}, $\pi_{0.5}$~\cite{DBLP:conf/corl/PhysicalIntelligence25}, SmolVLA~\cite{DBLP:preprint/arxiv/2506-01844}, GR00T-N1.5~\cite{DBLP:preprint/arxiv/2503-14734}, RTC~\cite{DBLP:conf/nips/BlackGL25}, VLA-Adapter-Pro~\cite{DBLP:preprint/arxiv/2509-09372}, and VLASH~\cite{DBLP:preprint/arxiv/2512-01031}, covering general-purpose VLAs, lightweight adaptation-based models, and latency-aware designs.
In real-world experiments, we evaluate $\pi_{0.5}$, SmolVLA, and VLASH under identical physical setups.
All baselines are initialized from publicly available pretrained weights and adapted to \ODslug using the same training data, following official recipes when available.

\noindent \textbf{Implementation Details.}
\OMplain is trained on 32 NVIDIA A100 GPUs with a batch size of 40 per GPU.
We use the AdamW optimizer with a learning rate of $1\times10^{-4}$, $\beta$ coefficients $(0.9, 0.95)$, $\epsilon=1\times10^{-8}$, and weight decay of $1\times10^{-10}$.
A cosine learning rate schedule with $1000$ warm-up steps is employed.
The models are trained for approximately two weeks, with three stages as 2 days for pre-training, 10 days for mid-training, and 2 days for post-training.
\OMplain requires 1.8GB of GPU memory and performs asynchronous action-chunk inference at 0.226 second per chunk on an NVIDIA RTX A6000.
Actions are indexed at the 25 Hz observation timestep with a fixed action horizon of $n = 20$, and the inference latency $m$ is measured at runtime and allowed to vary across hardware and execution cycles (Sec.~\ref{sec:exp-ablation}).

\noindent \textbf{Evaluation Metrics.}
We evaluate all methods using two metrics: 
\textit{Success Rate}, the fraction of trials that complete the instructed manipulation without object drop or timeout, and 
\textit{Task Completion Time}, the elapsed time from the onset of object motion to task termination, including successful completion, timeout, or object drop. 
We report both metrics in simulation and only success rate in real-world experiments. 
All metrics are averaged over multiple trials.
%

\noindent \textbf{Execution Constraints.}
To ensure safe real-world operation, we restrict the robot workspace within predefined bounds.
If the predicted end-effector position exceeds a predefined safety threshold, the robot aborts the current attempt and returns to a safe home pose, and the trial is marked as failure.

\subsection{Dynamic Interaction and Reactivity}
\label{sec:exp-interaction}

For the \emph{Interaction} dimension, the performance primarily depends on timely correction to evolving object motion.
Across CR/DA/LS (Table~\ref{tab:benchmark}), prior VLAs exhibit consistently low success under dynamic motion, while \OMplain reaches 60.5/38.5/40.5\%.
The gains are especially pronounced in long-horizon sequencing, where \OMplain improves over the strongest baseline by 33.0 percentage points.
This suggests that the main difficulty is not reacting once to a moving object, but maintaining temporal coordination as the object state continues to evolve during execution.

\begin{figure*}[!t]
  \resizebox{\linewidth}{!}{
    \includegraphics[width=\textwidth]{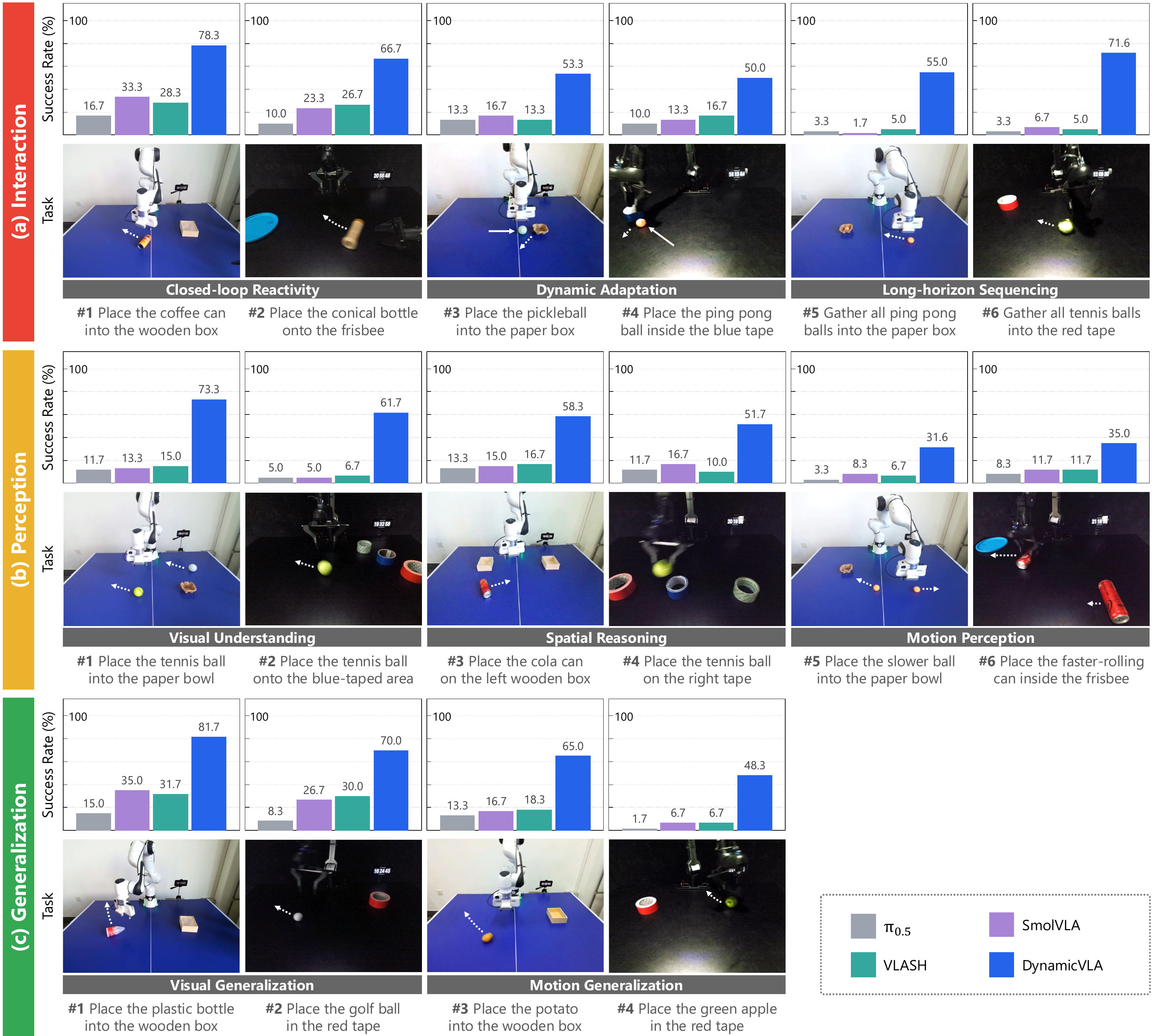}
  }
  \caption{\textbf{Real-world Evaluation.} We compare VLA models on 16 real-world dynamic manipulation tasks using Franka and PiPER as the manipulation robot. Each method--task pair is evaluated over 60 trials, comprising 20 trials for each of three paired motion--position configurations. An additional robot arm initializes object motion, providing comparable motion conditions across trials.}
  \label{fig:exp-real-world}
\end{figure*}

The same pattern appears in real-world trials (Figure~\ref{fig:exp-real-world}\hyperref[fig:exp-real-world]{a}). 
Baselines often commit to actions based on outdated observations or lose coordination after trajectory changes, whereas \OMplain more reliably maintains perception--execution alignment under tight timing constraints.

\begin{table*}[!t]
  \setlength\extrarowheight{1 pt}
  \caption{\textbf{\OD Simulation Benchmark Results.}  
  Average success rates (\%) are reported across nine evaluation sub-dimensions, organized under three categories: Interaction, Perception, and Generalization.
  In addition, overall average success rate (Success, \%) and task completion time (Time, seconds) are reported.
  Each method is evaluated over 1,800 trials (10 scenes $\times$ 9 dimensions $\times$ 20 trials). 
  All baseline models are fine-tuned on the DOM dataset using their official implementations and released pretrained weights.
  Best results are highlighted in bold.}
  \label{tab:benchmark}
  \resizebox{\linewidth}{!}{
    \begin{tabular}{l ccc ccc ccc ccc}
      \toprule
        \multirow{2}{*}{Methods} &  
        \multicolumn{3}{c}{\bf{Interaction}} &
        \multicolumn{3}{c}{\bf{Perception}} &
        \multicolumn{3}{c}{\bf{Generalization}} &
        \multicolumn{2}{c}{\bf{Average}} \\
       \cmidrule(lr){2-4} \cmidrule(lr){5-7} \cmidrule(lr){8-10} \cmidrule(lr){11-12}
       & CR        & DA         & LS
       & VU        & SR         & MP
       & VG        & MG         & DR
       & Success \upar & Time \dwar \\
      \midrule
        Diffusion Policy~\cite{DBLP:conf/rss/ChiFDXCBS23} &
        0.50        & 0.50      & 0.00  &
        1.00        & 0.00      & 0.00  &
        1.00        & 0.50      & 0.00  &
        0.38        & 10.89 \\
        ACT~\cite{DBLP:conf/rss/ZhaoKLF23} &
        2.50        & 1.50      & 0.00  &
        1.00        & 1.50      & 0.50  &
        2.50        & 2.50      & 0.00  &
        1.33        & 10.80 \\
        OpenVLA-OFT~\cite{DBLP:conf/rss/KimFL25} &
        3.50        & 0.50       & 0.50  &
        0.00        & 1.50       & 0.50  &
        3.50        & 2.00       & 0.00  &
        1.33        & 10.83 \\
        $\pi_{0.5}$~\cite{DBLP:conf/corl/PhysicalIntelligence25} & 
        9.50        & 17.50      & 3.50  &
        5.00        & 12.50      & 9.00  &
        5.00        & 19.50      & 18.00 &
        11.06       & 10.62 \\
        RTC~\cite{DBLP:conf/nips/BlackGL25} &
        12.00       & 15.00      & 4.00  &
        7.50        & 13.50      & 8.50  &
        6.50        & 14.00      & 20.00 &
        11.22       & 10.63 \\
        SmolVLA~\cite{DBLP:preprint/arxiv/2506-01844} &
        18.50       & 17.50      & 5.50  &
        1.50        & 14.50      & 11.50 &
        14.50       & 13.50      & 17.00 &
        12.67       & 10.65 \\
        GR00T-N1.5~\cite{DBLP:preprint/arxiv/2503-14734} &
        10.50       & 12.00      & 4.00  &
        9.50        & 13.50      & 14.00 &
        14.50       & 19.50      & 20.00 &
        13.05       & 10.56 \\
        VLA-Adapter-Pro~\cite{DBLP:preprint/arxiv/2509-09372} & 
        21.00       & 15.50      & 6.00  &
        6.50        & 16.50      & 10.50 &
        15.00       & 18.50      & 13.00 &
        13.61       & 9.98 \\
        VLASH~\cite{DBLP:preprint/arxiv/2512-01031} & 
        9.00        & 20.50      & 7.50  &
        6.50        & 7.50       & 12.00 &
        7.00        & 21.00      & 20.00 &
        12.33       & 10.60 \\
      \midrule
        \textbf{\OM} & 
        \bf{60.50}  & \bf{38.50} & \bf{40.50} &
        \bf{51.50}  & \bf{48.00} & \bf{33.50} &
        \bf{59.50}  & \bf{65.00} & \bf{26.50} & 
        \bf{47.06}  & \bf{8.53} \\
      \bottomrule
      \multicolumn{12}{l}{\small \makecell[l]{CR: Closed-loop Reactivity; DA: Dynamic Adaptation; LS: Long-horizon Sequencing; VU: Visual Understanding; SR: Spatial Reasoning;\\ MP: Motion Perception; VG: Visual Generalization; MG: Motion Generalization; DR: Disturbance Robustness}}
    \end{tabular}
  }
\end{table*}

\subsection{Multimodal Spatial-Temporal Reasoning}
\label{sec:exp-perception}

For the \emph{Perception} dimension, visual-language grounding must be performed under object motion. 
In simulation (Table~\ref{tab:benchmark}, VU/SR/MP), prior VLAs degrade substantially across all three sub-dimensions, while \OMplain improves success to 51.5/48.0/33.5\%. 
Perception-heavy dynamic tasks expose a trade-off between semantic grounding and real-time execution: policies must resolve appearance, spatial, and motion cues while acting before the scene evolves. 
Motion perception remains the most difficult case, suggesting that dynamic VLA policies still struggle when language grounding depends directly on object trajectories.

Real-world results show a similar trend (Figure~\ref{fig:exp-real-world}\hyperref[fig:exp-real-world]{b}).
\OMplain maintains a large advantage in visual and spatial grounding tasks, while motion-based instructions remain challenging, reflecting difficulty of estimating object motion, grounding language cues, and executing timely corrections.

\subsection{Generalization under Distribution Shift}
\label{sec:exp-generalization}

For the \emph{Generalization} dimension, we evaluate whether policies remain effective under distribution shifts in object appearance, scene layout, motion patterns, and disturbances.
In simulation (Table~\ref{tab:benchmark}, VG/MG/DR), \OMplain achieves 59.5/65.0/26.5\% success, consistently outperforming prior VLAs across all three settings.
These results indicate that \OMplain does not merely fit the training configurations, but retains its advantage under unseen objects, scenes, and motion regimes.

Disturbance robustness remains substantially more challenging, with a lower success rate than visual and motion generalization. 
This suggests that abrupt perturbations and contact changes still pose failure cases for current dynamic VLA policies. 
Real-world results show similar trends for visual and motion shifts (Figure~\ref{fig:exp-real-world}\hyperref[fig:exp-real-world]{c}).
We omit real-world disturbance trials because perturbation strength and contact conditions are difficult to reproduce reliably across methods.

\begin{table}[!t]
\begin{minipage}{.48\textwidth}
  \caption{\textbf{Ablation of key design choices.} Effects of FastViT vision encoding (FViT), Continuous Inference (CI), and Latency-aware Action Streaming (LAAS) on success rate (SR) on the \ODslug benchmark.}
  \vspace{1.9 mm}
  \label{tab:ablation-study}
  \begin{tabularx}{\linewidth}{lYYYc}
    \toprule
              & FViT    & CI      & LAAS   & SR (\%) \upar \\
    \midrule
       {[1]}  & \cmark  & \xmark  & \xmark & 30.27 \\
       {[2]}  & \cmark  & \xmark  & \cmark & 36.11 \\
       {[3]}  & \cmark  & \cmark  & \xmark & 39.72 \\
       {[4]}  & \xmark  & \cmark  & \cmark & 28.89 \\
    \midrule
       {[5]}  & \cmark  & \cmark  & \cmark & \bf{47.06} \\
    \bottomrule
  \end{tabularx}  
\end{minipage}
\hfill
\begin{minipage}{.48\textwidth}
  \caption{\textbf{Alternative overlap-handling strategies.} Naive async and Time Ensemble use the trained \OMplain policy \ul{with LAAS disabled}, while RTC uses a separately trained inpainting policy. All methods are evaluated under the same observation, action, and evaluation protocol. We report jerk and success rate (SR) on the \ODslug benchmark.}
  \label{tab:async-strategies}
  \begin{tabularx}{\linewidth}{lYc}
    \toprule
      Strategy      & Jerk \dwar & SR (\%) \upar \\
    \midrule
      Naive async   & 
      165.22        & 39.72 \\
      Time Ensemble~\cite{DBLP:conf/rss/ZhaoKLF23} & 
      160.69        & 39.72 \\
      RTC~\cite{DBLP:conf/nips/BlackGL25} & 
      \bf{155.36}   & \bf{39.83} \\
    \bottomrule
  \end{tabularx}
\end{minipage}
\end{table}

\subsection{Ablation Studies}
\label{sec:exp-ablation}

To analyze \OMplain's main design choices, we study visual encoding, continuous inference and overlap handling, LAAS, and the action horizon–latency relation.
All variants are evaluated on \ODslug under identical protocols, with results summarized in Tables~\ref{tab:ablation-study} and~\ref{tab:async-strategies}.

\noindent \textbf{Vision Encoding.}
We ablate the visual encoder by replacing FastViT with a transformer-based encoder following the SmolVLM configuration~\cite{DBLP:preprint/arxiv/2504-05299}, while keeping all other components fixed.
Measured on a single NVIDIA RTX A6000 GPU, the FastViT-based variant reduces end-to-end per-chunk inference time from 0.523s to 0.226s and improves success from 28.89\% to 47.06\% (Table~\ref{tab:ablation-study}, [4] and [5]).
This supports the use of efficient visual encoding for multi-view, multi-frame dynamic manipulation under tight latency constraints.

\noindent \textbf{Continuous Inference and Overlap Handling.}
We evaluate the role of the continuous inference schedule by disabling it while keeping all other components unchanged (Table~\ref{tab:ablation-study}, [2] and [5]).
CI uses naive asynchronous execution, launching each inference call immediately after the previous one finishes and selecting latest-chunk actions without smoothing.
Disabling CI introduces inter-chunk waiting and reduces success from 47.06\% to 36.11\%.
As shown in Table~\ref{tab:async-strategies}, Time Ensemble and RTC strategies slightly reduce jerk, measured as the time derivative of acceleration, but leave success nearly unchanged relative to naive async execution, remaining below the full LAAS variant in Table~\ref{tab:ablation-study} [5].
This suggests that learned chunks are already temporally consistent enough for naive async execution, while overlap smoothing alone does not address latency-invalid actions under dynamic object motion.

\noindent \textbf{Latency-aware Action Streaming.}
We evaluate LAAS as a mechanism for correcting the perception--execution mismatch caused by inference latency.
As shown in Table~\ref{tab:ablation-study}, removing latency-invalid prefixes improves success even without Continuous Inference ([1] vs. [2]: 30.27\% to 36.11\%), indicating that actions indexed to already elapsed timesteps can hurt execution.
Under Continuous Inference, disabling LAAS reduces success from 47.06\% to 39.72\% ([5] vs. [3]), showing that non-blocking execution alone does not remove this temporal misalignment.
These results suggest that CI and LAAS address complementary latency effects: CI reduces inter-chunk waiting, while LAAS removes stale prefixes to preserve action-time alignment under dynamic object motion.

\noindent \textbf{Action Horizon $n$ and Inference Latency $m$.}
Actions are indexed at the 25 Hz observation timestep, so $n = 20$ spans 0.8 s and the 0.226 s per-chunk latency corresponds to $m \approx 6$ steps.
LAAS uses the online-measured latency to discard a variable-length latency-invalid prefix from each completed chunk.
Since $n \ge 2m$ on our hardware, the remaining suffix of length $n - m$ still covers the next inference delay, avoiding inter-chunk waiting.
A sensitivity study in Appendix~\ref{sec:discussion} (Table~\ref{tab:variable-latency}) injects $J$ extra steps of delay into the same policy: success is unchanged while $n \ge 2(m + J)$ holds and degrades once the valid suffix no longer covers the next inference call.
This also delimits the scope of CI and LAAS: their benefit requires a backbone whose latency satisfies the non-blocking condition, and each executed action still rests on observations $m$ steps old, which is why integrating them into existing VLAs (Appendix~\ref{sec:discussion}) yields a clear gain for SmolVLA but only a limited gain for the higher-latency $\pi_{0.5}$.

\section{Conclusion and Future Work}

This work shows that dynamic object manipulation exposes a key failure mode for VLA models: latency-induced temporal misalignment between observation and action execution, which is largely hidden in static manipulation. 
We present \OM, a latency-aware VLA model that combines efficient multimodal inference, non-blocking continuous inference, and Latency-aware Action Streaming to discard latency-invalid prefixes and preserve action-time alignment. 
We further introduce DOM, a large-scale benchmark built from scalable simulation and real-world data collection pipelines with automated demonstrators.
Looking forward, several limitations of the current study point to promising directions for future work:

\noindent \textbf{Toward Latency-scalable VLA Architectures.}
Dynamic manipulation exposes a fundamental tension between semantic understanding and real-time responsiveness.
Future VLAs should improve inference speed without simply sacrificing visual-language reasoning, through adaptive token selection, compact temporal context, and efficient action generation.

\noindent \textbf{Toward Longer-horizon Dynamic Tasks.}
\ODslug evaluates long-horizon sequencing through sustained interaction with moving objects.
Future dynamic tasks can further extend to multi-stage settings where goals, object motion, and contact outcomes evolve over longer horizons, requiring memory and planning under real-time constraints.

\noindent \textbf{Toward Richer Physical Dynamics.}
Moving rigid objects provide a tractable setting for studying perception--execution alignment.
Future work can extend dynamic manipulation to deformable, articulated, granular, or fluid objects, where state prediction becomes more challenging.

\section*{Acknowledgments}

%
We thank Prof. David Hsu (NUS) and Prof. Shengping Zhang (HIT) for their support and for providing access to the Franka Emika Panda robot utilized in this work.
This work was supported by the Singapore Ministry of Education under its Academic Research Fund Tier 2 (MOE-T2EP20221-0012, MOE-T2EP20223-0002), and by cash and in-kind contributions from NTU S-Lab and industry partner(s).

\clearpage
\bibliographystyle{IEEEtran}
\bibliography{references}

\clearpage
\appendix
\section{Model Architecture Details}

\noindent \textbf{VLM Backbone.}
The RGB images in the temporal observation window $\mathbf{O}_t$  are concatenated and encoded by FastViT~\cite{DBLP:conf/iccv/VasuGZTR23} using a hierarchical multi-stage design. 
Each input image is resized to $384 \times 384$, and the encoder progressively increases channel width across stages $(96, 192, 384, 768, 1536)$ with corresponding block depths $(2, 12, 24, 4, 2)$.
FastViT applies aggressive spatial compression via a large initial patch size of $64$ and strided downsampling, using RepMixer-style token mixing in early stages and attention in later ones. 
The encoder outputs $36$ visual tokens of fixed dimension $960$, aligned with the language embedding space, achieving substantial token reduction while preserving manipulation-relevant spatial structure.
In addition to visual inputs, the robot proprioceptive state $\mathbf{P}_t$ is incorporated as an explicit conditioning signal. 
The $32$-dimensional state vector, containing Cartesian position and orientation with zero padding for unused entries, is linearly projected into the language embedding space and represented as a single $960$-dimensional state token.
Language instructions $\mathbf{L}_t$ are tokenized into a variable number of language tokens depending on prompt length. 
All visual, language, and state tokens are concatenated and processed jointly by the language backbone.
Multimodal reasoning is performed by SmolLM2-360M~\cite{DBLP:preprint/arxiv/2504-05299}, where only the first $16$ transformer layers are used to reduce inference latency, following the practice adopted in SmolVLA~\cite{DBLP:preprint/arxiv/2506-01844}.
The backbone outputs key–value representations for all processed tokens, which are cached and reused across inference cycles. 

\noindent \textbf{Action Expert.}
Action generation is handled by a dedicated diffusion-based action expert, instantiated as a lightweight transformer copied from the language backbone and truncated to the first 16 layers. 
The expert predicts an action chunk with horizon $n = 20$, which is sufficient under Continuous Inference while keeping inference latency low. 
Each action is a $32$-dimensional vector representing end-effector pose and gripper state (with zero padding), and the noisy action input has shape $(n, 32)$ during training and pure noise during inference.
The action expert uses a reduced hidden dimension of $720$ (0.75 $\times$ the language embedding size) to lower computation. 
Noisy action tokens are projected into this space and combined with diffusion timestep embeddings, and denoising updates are generated by querying the cached key–value representations, without re-encoding perceptual inputs.

\section{The Training Scheme}

\noindent \textbf{Pre-training Stage.}
The vision–language backbone combines a convolutional visual encoder (FastViT) and a compact language model (SmolLM2-360M), both initialized from their respective pretrained weights.
To align visual and linguistic representations, we first perform large-scale vision–language pre-training using 150M English image–text pairs sampled from COYO-700M~\cite{DBLP:preprint/github/coyo700m}.

\noindent \textbf{Mid-training Stage.}
After vision–language pre-training, the full VLA model is trained on the synthetic \OD (\ODslug) dataset (Sec.~\ref{sec:dom-dataset}). 
Each episode provides temporally evolving multi-view visual observations, from which the model uses a wrist-mounted camera on the end-effector and a fixed third-person camera facing the manipulator.
To capture short-term dynamics, the temporal observation window is instantiated as $\mathbf{O}_t = \{\mathbf{o}_{t - 2}, \mathbf{o}_t\}$.
Using two views per timestep, this results in four images per input step, which are concatenated channel-wise and processed jointly by the vision encoder.
In this stage, \OMplain is optimized using minibatches formed by randomly sampling episode timesteps from shuffled manipulation demonstrations. 
For each minibatch, the model is trained on tuples $(\mathbf{O}_t, \mathbf{L}_t, \mathbf{P}_t)$, while the action expert is trained to denoise a noisy action chunk $\mathbf{A}_t^\tau$ under the objective defined in Eq.~\ref{eq:flow-matching-loss}.

\noindent \textbf{Post-training stage.}
In the post-training stage, the model is fine-tuned on robot-specific real-world demonstrations using the same objective as in mid-training, enabling adaptation to new embodiments and sensing configurations.

\section{More Discussion}
\label{sec:discussion}

\begin{table}[!t]
  \centering
  \caption{\textbf{Ablation on Temporal Visual Context.} The temporal observation window is varied by enabling different visual frames at time steps $\{t-3, t-2, t-1, t\}$, while keeping the model architecture, inference frequency, and execution pipeline fixed. Note that SR, T.Time, and I.Time represent the success rate (in \%), task completion time (in seconds), and inference time (in seconds, measured on an NVIDIA RTX A6000 GPU), respectively.}
  \label{tab:temporal-visual-context}
  \begin{tabularx}{\linewidth}{YYYYYYY}
    \toprule
       $t - 3$    & $t - 2$    & $t - 1$      & $t$ &
       SR \upar   & T.Time \dwar & I.Time \dwar \\
    \midrule
       \xmark     & \xmark     & \xmark       & \cmark     & 
       38.22      & 9.52       & \bf{0.225} \\
       \xmark     & \xmark     & \cmark       & \cmark     &
       43.39      & 8.77       & 0.226 \\
       \xmark     & \cmark     & \xmark       & \cmark     &
       47.06      & 8.53       & 0.226 \\
       \cmark     & \xmark     & \xmark       & \cmark     &
       46.89      & 8.51       & 0.226 \\
       \xmark     & \cmark     & \cmark       & \cmark     &
       \bf{47.11} & \bf{8.46}  & 0.228 \\
       \cmark     & \cmark     & \cmark       & \cmark     &
       47.06      & 8.53       & 0.229 \\
    \bottomrule
  \end{tabularx}
\end{table}

\noindent \textbf{Temporal Visual Context.}
We conduct an ablation study to analyze the impact of temporal visual context by varying the composition of the observation window $\mathbf{O}_t$ within the same \OMplain architecture. 
As described in Sec.~\ref{sec:method}, our default setting feeds the model $\mathcal{M}$ with a sparse temporal window $\mathbf{O}_t = \{\mathbf{o}_{t-2}, \mathbf{o}_t\}$, which is designed to facilitate implicit object velocity perception.
As summarized in Table~\ref{tab:temporal-visual-context}, different temporal configurations lead to negligible differences in inference latency and parameter count. 
We observe that using a single-frame input $\{\mathbf{o}_t\}$ results in a clear drop in task success rate, as a single observation lacks the temporal cues necessary for estimating object motion and dynamics. 
However, expanding the temporal window beyond two frames does not yield further noticeable gains, indicating diminishing returns from additional visual redundancy. 
Moreover, compared to $\{\mathbf{o}_{t-2}, \mathbf{o}_t\}$, the setting $\{\mathbf{o}_{t-1}, \mathbf{o}_t\}$ achieves lower success rates, suggesting that a larger temporal interval provides more informative motion cues for velocity estimation. 
Overall, these results demonstrate that sparse but sufficiently spaced temporal context is critical for effective dynamic manipulation, even without increasing inference frequency.

\begin{table}[!t]
  \centering
  \caption{\textbf{Ablation on LLM Depth.} Different LLM depths are evaluated by retaining the first $l$ transformer layers. Note that SR, T.Time, I.Time, and \#Param denote success rate (\%), task completion time (seconds), inference time (seconds, measured on an NVIDIA RTX A6000 GPU), and parameter count (in millions), respectively.}
  \label{tab:llm-depth}
  \begin{tabularx}{\linewidth}{YYYYc}
    \toprule
       \#Layers   &
       SR \upar   & T.Time \dwar & I.Time \dwar & \#Param \dwar \\
    \midrule
       8          & 
       44.17      & 8.92         & \bf{0.127}   & \bf{303} \\
       16         & 
       47.06      & 8.53         & 0.226        & 430 \\
       24         & 
       \bf{48.44} & 8.43         & 0.317        & 558 \\
       32         & 
       42.11      & \bf{8.39}    & 0.373        & 685 \\
    \bottomrule
  \end{tabularx}
\end{table}

\noindent \textbf{Depth of LLM Backbone.}
Following the backbone truncation strategy~\cite{DBLP:preprint/arxiv/2506-01844}, we reduce inference latency by retaining only the first $l$ transformer layers of the LLM during inference.
To examine whether this design choice remains effective in the \OMplain setting, we evaluate multiple backbone depths ($l=8, 16, 24$) and compare them against the full model ($l=32$).
As shown in Table~\ref{tab:llm-depth}, increasing the backbone depth leads to a modest increase in inference latency.
However, this additional latency can be largely amortized by Continuous Inference and Latency-aware Action Streaming, and does not translate into a noticeable improvement in task success rate.
In contrast, aggressively truncating the backbone significantly improves inference speed, but at the cost of reduced model capacity, resulting in a substantial degradation in success rate.
Overall, this ablation confirms that a 16-layer backbone strikes the optimal balance between efficiency and robustness.

\begin{table}[!t]
\begin{minipage}{.48\textwidth}
  \caption{\textbf{Cross-Model Analysis of CI and LAAS.} CI and LAAS are integrated into existing VLA models without backbone modification or retraining. 
  Note that SR and T.Time represent the success rate (in \%) and task completion time (in seconds), respectively.
  $^\dag$ indicates inference-time integration of CI and LAAS.}
  \label{tab:cross-model-ci-laas}
  \centering
  \begin{tabularx}{\linewidth}{lYc}
    \toprule
       Method        & 
       SR (\%) \upar & T.Time (s) \dwar \\
    \midrule
       $\pi_{0.5}$$^\dag$~\cite{DBLP:conf/corl/PhysicalIntelligence25} & 
       15.89         & 9.95\\
       SmolVLA$^\dag$~\cite{DBLP:preprint/arxiv/2506-01844} & 
       25.56         & 9.77 \\
    \midrule
       \OM &
       \bf{47.06}    & \bf{8.53} \\
    \bottomrule
  \end{tabularx}

\end{minipage}
\hfill
\begin{minipage}{.48\textwidth}
  \caption{\textbf{Robustness to increased inference latency.} $J$ denotes the number of extra control steps inserted after each inference call before the predicted action chunk is released. Success rate (SR) and task completion time (T.Time) are reported on the \ODslug benchmark.}
  \label{tab:variable-latency}
  \vspace{2.35 mm}
  \begin{tabularx}{\linewidth}{YYY}
    \toprule
       Delay $J$ & SR (\%) \upar & T.Time \dwar \\
    \midrule
       2         & \bf{47.06}    & \bf{8.53} \\
       4         & \bf{47.06}    & \bf{8.53} \\
       6         & 44.44         & 8.79 \\
       8         & 41.94         & 8.97 \\
    \bottomrule
  \end{tabularx}

\end{minipage}
\end{table}

\noindent \textbf{Cross-Model Analysis of CI and LAAS}
To evaluate the generality of the proposed execution mechanisms, Continuous Inference (CI) and Latency-aware Action Streaming (LAAS) are integrated into existing VLA models, including SmolVLA and $\pi_{0.5}$, without altering their backbone architectures.
As shown in Table~\ref{tab:cross-model-ci-laas}, consistent performance improvements are observed on SmolVLA, indicating that CI and LAAS effectively enhance closed-loop responsiveness under moderate inference latency.
In contrast, $\pi_{0.5}$ exhibits only marginal gains, as its substantially larger backbone incurs high inference latency, which limits the effectiveness of overlapping inference and temporally aligned execution.
These results suggest that CI and LAAS are broadly applicable, while their practical benefit is maximized when paired with a low-latency backbone.

\noindent \textbf{Sensitivity to Additional Inference Latency.}
Although the main formulation uses a constant $m$ for notation, our implementation measures latency online for each inference call.
When a chunk launched at time $t$ becomes available after $\hat{m}_t$ control steps, LAAS discards the first $\hat{m}_t$ actions and streams the remaining valid suffix.
To test robustness to increased inference latency, we inject a fixed extra delay after each inference call. 
Specifically, before releasing each predicted chunk, we wait for an additional $J$ control steps, so the effective latency becomes $\hat{m}_t = m_t + J$. 
We evaluate $J \in \{2, 4, 6, 8\}$ under the same DOM protocol.
On a single NVIDIA RTX A6000 GPU, the nominal latency $m_t \approx 6$ control steps. 
With $n = 20$, the executable suffix remains long enough to bridge the next inference call up to approximately $J = 4$ to satisfy $n \ge 2\hat{m}_t$. 
Once the effective latency grows further, the suffix length $n - m_t$ becomes insufficient to cover the next inference interval, leading to inter-chunk waiting and lower success rates.
As shown in Table~\ref{tab:variable-latency}, performance remains stable under moderate additional delay and degrades only when the non-blocking condition is violated.

\section{Detailed Evaluation Setup}
\label{sec:eval-setup}

In this section, we provide comprehensive details of the simulation data split and the real-world evaluation setups used in our experiments, including task specifications and object configurations.

\noindent \textbf{Simulation Data Split and Evaluation Protocol.}
The training pool consists of 201 of the 206 objects and 2,821 of the 2,824 scenes, with object speeds from 0.15 to 0.75 m/s plus static objects and friction coefficients from 0.5 to 1.5.
Language instructions are generated from attribute tags (appearance, size, spatial position, distance, and speed) ranked by per-episode object states, and test instructions use held-out object--target combinations.
\emph{Visual Generalization} evaluates on 5 held-out objects and 3 held-out scenes that are disjoint from the training pool by asset ID.
\emph{Motion Generalization} evaluates on out-of-range speeds (0.75 to 1.0 m/s) and out-of-range friction coefficients (0 to 0.5 and 1.5 to 2.0).
\emph{Disturbance Robustness} applies scripted pushes and collisions at evaluation time only.
The remaining six sub-dimensions use training-distribution objects and scenes with unseen per-trial initial states.
Each sub-dimension uses 10 test environments with 20 trials each.
An environment specifies a scene asset, its objects, and pre-generated initial object poses, velocities, and friction values that are reused identically across all methods, and disturbance events are scripted per trial.
No object or scene asset ID appears in both a training set and its held-out evaluation set; the exact ID lists and per-trial specifications ship with the benchmark release.

\noindent \textbf{Real-world Object Launching.}
Each real-world task is executed under standardized conditions to ensure repeatability and fair comparison across different policies.
Objects are launched by a secondary robot arm that executes a position-controlled push-and-release primitive with a trapezoidal velocity profile (maximum acceleration 2.0 m/s$^2$), releasing the object 2 cm above the tabletop.
Release speeds range from 0.2 to 0.8 m/s and are verified post hoc by the tracking system.
Evaluation is conducted across three predefined paired motion--position configurations, each combining an initial motion profile with a corresponding target container position; the three configurations use release speeds of 0.3, 0.5, and 0.7 m/s.
Trajectory families cover straight rolls, curved rolls (rotationally asymmetric objects such as the conical bottle), and collision-deflected trajectories.
Across repeated launches of the same configuration, the standard deviation is 0.03 m/s in release speed and 3.5 cm in trajectory endpoint, confirming comparable motion conditions across methods.

\needspace{5\baselineskip}
\noindent \textbf{Real-world Interaction Evaluation} (Sec.~\ref{sec:exp-interaction})
\begin{itemize}
\item
\textbf{Place the coffee can into the wooden box.}
The robot must track and grasp a rolling Nescaf{\'e} coffee can and place it into a wooden box.
This task evaluates closed-loop reactivity to continuously moving targets.

\item
\textbf{Place the conical bottle onto the frisbee.}
The robot must grasp a conical roasted sesame bottle whose rolling motion follows a curved trajectory and place it onto a blue frisbee.
This task evaluates closed-loop reactivity under non-linear object motion.

\item
\textbf{Place the pickleball into the paper box.}
The robot must grasp a moving pickleball and place it into a paper box, where the ball is designed to collide with the box and undergo trajectory deflection.
This task evaluates adaptive manipulation under contact-induced motion changes.

\item
\textbf{Place the ping pong ball inside the blue tape.}
The robot must grasp a moving ping pong ball and place it within a blue-taped region, where impacts with the tape are designed to deflect the ball's trajectory.
This task evaluates adaptive placement under perturbed object motion.

\item
\textbf{Gather all ping pong balls into the paper box.}
The robot must continuously collect ping pong balls that repeatedly appear on the tabletop and place them into a paper box.
This task evaluates long-horizon task sequencing under sustained dynamic inputs.

\item
\textbf{Gather all tennis balls into the red tape.}
The robot must continuously collect tennis balls that repeatedly appear on the tabletop and return them to a red-taped region.
This task evaluates long-horizon planning and execution in dynamic environments.
\end{itemize}

\needspace{5\baselineskip}
\noindent \textbf{Real-world Perception Evaluation} (Sec.~\ref{sec:exp-perception})
\begin{itemize}
\item
\textbf{Place the tennis ball into the paper bowl.}
The robot must identify and grasp the moving tennis ball among multiple simultaneously thrown objects (a tennis ball and a pickleball), and place it into a paper bowl.
This task evaluates object-level visual understanding for identifying and manipulating the correct target under dynamic motion.

\item
\textbf{Place the tennis ball onto the blue-taped area.}
The robot is required to catch a rolling tennis ball and place it precisely within the region marked by blue tape, among multiple visually similar tape markings (red, blue, and transparent).
This task evaluates visually grounded target understanding and precise placement under continuous object motion.

\item
\textbf{Place the cola can on the left wooden box.}
The robot must grasp a moving cola can and place it on a wooden box located to its left, evaluating its ability to handle spatial placement under varying object motions.
This task evaluates spatial understanding for target localization and placement relative to the robot’s viewpoint.

\item
\textbf{Place the tennis ball on the right tape.}
The robot must grasp a moving tennis ball and place it on a tape located to its right, evaluating spatial awareness and placement precision.
This task evaluates spatial understanding for interpreting directionally specified targets and executing accurate placement.

\item
\textbf{Place the slower ball into the paper bowl.}
The robot must grasp the ping pong ball specified by its lower moving speed and place it into the paper bowl.
This task evaluates motion-based target understanding, where the target is specified by its movement direction.

\item
\textbf{Place the faster-rolling can inside the frisbee.}
The robot must grasp the cola can specified by its higher rolling speed and place it inside the blue frisbee.
This task evaluates motion-based target understanding, where the target is specified by its relative motion speed.
\end{itemize}

\needspace{5\baselineskip}
\noindent \textbf{Real-world Generalization Evaluation} (Sec.~\ref{sec:exp-generalization})
\begin{itemize}
\item
\textbf{Place the plastic bottle into the wooden box.}
The robot must grasp a rolling plastic bottle with an unseen appearance and a regular curved trajectory, and place it into a wooden box.
This task evaluates visual generalization to unseen object appearances under dynamic motion.

\item
\textbf{Place the golf ball in the red tape.}
The robot must grasp a rolling golf ball with an unseen appearance and place it within a red-taped region.
This task evaluates visual generalization to unseen object instances during dynamic manipulation.

\item
\textbf{Place the potato into the wooden box.}
The robot must grasp a moving potato whose motion follows irregular patterns and place it into a wooden box.
This task evaluates motion generalization to irregular object dynamics.

\item
\textbf{Place the green apple in the red tape.}
The robot must grasp a moving green apple whose motion exhibits irregular and unpredictable patterns, and place it onto a red-taped region.
This task evaluates motion generalization to irregular object trajectories.
\end{itemize}

\section{Broader Impacts}

\OMplain may support research on more reliable robotic manipulation in settings with moving objects, such as logistics, manufacturing, and other structured service scenarios. 
By identifying latency-induced perception--execution mismatch and introducing DOM, this work may also help the community evaluate VLA policies under more realistic timing constraints. 
As with other physical robot learning systems, downstream deployment requires careful safety validation, especially in unstructured environments or around humans. 
Our experiments are conducted in controlled simulation and laboratory settings with bounded workspaces and abort conditions; the model is intended as a research prototype rather than an unsupervised system for safety-critical deployment.

\section{Existing Assets and Licenses}

We use several existing models, datasets, simulators, and perception tools. All external assets are used under their respective licenses or terms of use, and we cite original creators in the main paper.

\begin{itemize}[leftmargin=*]
\item \textbf{Pretrained models and tools.}
\OMplain uses SmolLM2 as the language backbone, FastViT as the visual encoder, and EfficientTAM for real-world object mask tracking. SmolLM2 and EfficientTAM are released under Apache-2.0, while FastViT is used under Apple’s ML-FastViT license terms.

\item \textbf{Pretraining and simulation data.} 
We use COYO-700M for vision-language pretraining, Objaverse objects for simulation assets, and 3D-FRONT scenes for simulated environments. COYO-700M is distributed under CC-BY-4.0, while its underlying image/text content remains subject to the licenses of the original content providers; Objaverse is distributed under ODC-By v1.0, with individual objects under their corresponding Creative Commons licenses; 3D-FRONT is used under its dataset terms of use.

\item \textbf{Simulation platform.} 
We use Isaac Sim for physics simulation and rendering. The Isaac Sim source code is released under Apache-2.0, while additional components such as Omniverse Kit SDK, models, and textures are governed by separate NVIDIA license terms.
\end{itemize}

\clearpage
\section*{NeurIPS Paper Checklist}

\begin{enumerate}

\item {\bf Claims}
    \item[] Question: Do the main claims made in the abstract and introduction accurately reflect the paper's contributions and scope?
    \item[] Answer: \answerYes{}
    \item[] Justification: The abstract and introduction state the main contributions: identifying latency-induced perception–execution mismatch, proposing DynamicVLA with continuous inference and LAAS, introducing DOM, and evaluating on simulation and real robots. These claims are directly supported by the method, benchmark, and experimental sections.

\item {\bf Limitations}
    \item[] Question: Does the paper discuss the limitations of the work performed by the authors?
    \item[] Answer: \answerYes{}
    \item[] Justification: The paper discusses limitations and future directions, including the latency–understanding trade-off, limited focus on short- to medium-horizon dynamics, and the current assumption of rigid-body dynamics. These points define the scope of the proposed method and indicate important cases not yet addressed.

\item {\bf Theory assumptions and proofs}
    \item[] Question: For each theoretical result, does the paper provide the full set of assumptions and a complete (and correct) proof?
    \item[] Answer: \answerNA{}
    \item[] Justification: The paper does not present formal theoretical results, theorems, or proof-based claims. The mathematical formulation is used to define the dynamic manipulation setting and proposed execution mechanism rather than to establish theoretical guarantees.

    \item {\bf Experimental result reproducibility}
    \item[] Question: Does the paper fully disclose all the information needed to reproduce the main experimental results of the paper to the extent that it affects the main claims and/or conclusions of the paper (regardless of whether the code and data are provided or not)?
    \item[] Answer: \answerYes{}
    \item[] Justification: The paper provides the model architecture, data-collection pipeline, benchmark dimensions, evaluation protocols, baselines, metrics, and implementation details needed to understand and reproduce the main experiments. The experimental setup specifies the simulation and real-robot platforms, trial counts, input modalities, baseline adaptation, and evaluation metrics.

\item {\bf Open access to data and code}
    \item[] Question: Does the paper provide open access to the data and code, with sufficient instructions to faithfully reproduce the main experimental results, as described in supplemental material?
    \item[] Answer: \answerYes{}
    \item[] Justification: The code and data are open-sourced on GitHub at \url{https://github.com/hzxie/DynamicVLA}, with instructions to reproduce the main results.

\item {\bf Experimental setting/details}
    \item[] Question: Does the paper specify all the training and test details (e.g., data splits, hyperparameters, how they were chosen, type of optimizer) necessary to understand the results?
    \item[] Answer: \answerYes{}
    \item[] Justification: The paper specifies the training stages, input construction, optimizer, learning rate, batch size, compute setup, inference configuration, action horizon, and evaluation protocol. The appendix further reports temporal-context and model-depth ablations, supporting the interpretation of the main results.

\item {\bf Experiment statistical significance}
    \item[] Question: Does the paper report error bars suitably and correctly defined or other appropriate information about the statistical significance of the experiments?
    \item[] Answer: \answerNo{}
    \item[] Justification: We report success rates and task completion times over fixed simulation and real-world trial sets, but do not include error bars or formal significance tests. Estimating variance over independent evaluation replications would require repeatedly resampling scenes, motion conditions, object states, and contact outcomes, which is beyond the scope of the current benchmark evaluation.

\item {\bf Experiments compute resources}
    \item[] Question: For each experiment, does the paper provide sufficient information on the computer resources (type of compute workers, memory, time of execution) needed to reproduce the experiments?
    \item[] Answer: \answerYes{}
    \item[] Justification: Section~5.1 reports the compute resources used for training and inference, including GPU type and count, per-GPU batch size, training time, inference GPU, memory usage, action timestep rate, and action-chunk inference latency. Baseline and ablation experiments follow the same compute setup unless otherwise specified.
    
\item {\bf Code of ethics}
    \item[] Question: Does the research conducted in the paper conform, in every respect, with the NeurIPS Code of Ethics \url{https://neurips.cc/public/EthicsGuidelines}?
    \item[] Answer: \answerYes{}
    \item[] Justification: The research is conducted on simulated and controlled real-robot manipulation tasks and includes safety constraints for real-world execution. In particular, the robot workspace is bounded, unsafe predicted end-effector positions trigger abort and return-to-home behavior, and failed trials are explicitly counted as failures.

\item {\bf Broader impacts}
    \item[] Question: Does the paper discuss both potential positive societal impacts and negative societal impacts of the work performed?
    \item[] Answer: \answerYes{}
    \item[] Justification: The paper discusses broader impacts in the appendix, including potential benefits for dynamic robotic manipulation and risks associated with deploying VLA policies on physical robots. It also clarifies that the experiments are conducted in controlled settings and that deployment in human-populated or safety-critical environments would require additional safety validation and safeguards.
    
\item {\bf Safeguards}
    \item[] Question: Does the paper describe safeguards that have been put in place for responsible release of data or models that have a high risk for misuse (e.g., pre-trained language models, image generators, or scraped datasets)?
    \item[] Answer: \answerNA{}
    \item[] Justification: The paper does not release high-risk generative models, scraped web-scale datasets, or dual-use models of the kind highlighted in the checklist. For physical robot experiments, the paper nevertheless describes operational safeguards such as workspace limits and abort-and-return behavior during unsafe predictions.

\item {\bf Licenses for existing assets}
    \item[] Question: Are the creators or original owners of assets (e.g., code, data, models), used in the paper, properly credited and are the license and terms of use explicitly mentioned and properly respected?
    \item[] Answer: \answerYes{}
    \item[] Justification: We cite the creators of all existing assets used in this work, including pretrained models, datasets, simulators, and tracking tools. Their licenses or terms of use are summarized in the appendix and release documentation.

\item {\bf New assets}
    \item[] Question: Are new assets introduced in the paper well documented and is the documentation provided alongside the assets?
    \item[] Answer: \answerYes{}
    \item[] Justification: The paper introduces the DOM benchmark, including synthetic and real-world episodes, simulation-ready scenes, object sets, benchmark dimensions, and evaluation tasks. The data-collection pipeline, object dynamics, sensors, task categories, and real-world evaluation tasks are documented in the main paper and appendix.

\item {\bf Crowdsourcing and research with human subjects}
    \item[] Question: For crowdsourcing experiments and research with human subjects, does the paper include the full text of instructions given to participants and screenshots, if applicable, as well as details about compensation (if any)? 
    \item[] Answer: \answerNA{}
    \item[] Justification: The paper does not involve crowdsourcing or human-subject experiments. Data are collected through simulation and automated robot data-collection pipelines rather than through human participants.

\item {\bf Institutional review board (IRB) approvals or equivalent for research with human subjects}
    \item[] Question: Does the paper describe potential risks incurred by study participants, whether such risks were disclosed to the subjects, and whether Institutional Review Board (IRB) approvals (or an equivalent approval/review based on the requirements of your country or institution) were obtained?
    \item[] Answer: \answerNA{}
    \item[] Justification: The work does not involve human subjects, crowdsourcing workers, or participant studies, so IRB approval is not applicable.

\item {\bf Declaration of LLM usage}
    \item[] Question: Does the paper describe the usage of LLMs if it is an important, original, or non-standard component of the core methods in this research? Note that if the LLM is used only for writing, editing, or formatting purposes and does \emph{not} impact the core methodology, scientific rigor, or originality of the research, declaration is not required.
    \item[] Answer: \answerYes{}
    \item[] Justification: The paper describes the use of SmolLM2-360M as an existing pretrained language backbone in \OMplain, using its first 16 transformer layers for efficient multimodal reasoning. The LLM is not newly developed in this work, but it is a core component of the proposed VLA architecture and is therefore declared.

\end{enumerate}

\end{document}